%% file: main.tex
\documentclass[conference]{IEEEtran}
\IEEEoverridecommandlockouts
\usepackage{cite}
\usepackage{amsmath,amssymb,amsfonts}
\usepackage{algorithmic}
\usepackage{graphicx}
\usepackage{textcomp}
\usepackage{xcolor}
\usepackage{subcaption}
\usepackage[colorlinks,urlcolor=blue,linkcolor=blue,citecolor=blue]{hyperref}
\usepackage{caption}
\usepackage{wrapfig}

\def\BibTeX{{\rm B\kern-.05em{\sc i\kern-.025em b}\kern-.08em
    T\kern-.1667em\lower.7ex\hbox{E}\kern-.125emX}}

\makeatletter 
\newcommand{\linebreakand}{%
  \end{@IEEEauthorhalign}
  \hfill\mbox{}\par
  \mbox{}\hfill\begin{@IEEEauthorhalign}
}
\makeatother 

\begin{document}

\title{Grounding from an AI and Cognitive Science Lens}

\author{\IEEEauthorblockN{Goonmeet Bajaj}
\IEEEauthorblockA{The Ohio State University \\
bajaj.32@osu.edu}
\and
\IEEEauthorblockN{Srinivasan Parthasarathy}
\IEEEauthorblockA{The Ohio State University \\
srini@cse.ohio-state.edu}
\and
\IEEEauthorblockN{Valerie L. Shalin}
\IEEEauthorblockA{Wright State University \\
valerie.shalin@wright.edu}
\and 
\linebreakand 
\IEEEauthorblockN{Amit Sheth}
\IEEEauthorblockA{University of South Carolina \\
amit@sc.edu}
}

\maketitle

\begin{abstract}
 Grounding is a challenging problem, requiring a formal definition and different levels of abstraction. This article explores grounding from both cognitive science and machine learning perspectives. It identifies the subtleties of grounding, its significance for collaborative agents, and similarities and differences in grounding approaches in both communities. The article examines the potential of neuro-symbolic approaches tailored for grounding tasks, showcasing how they can more comprehensively address grounding. Finally, we discuss areas for further exploration and development in grounding.
\end{abstract}


\section*{Introduction}
Robust communication transcends human-human communication settings to include human-machine, machine-machine, and multi-agent human-machine teams. 
\textit{Grounding} fosters a common understanding among agents performing a task - typically in the real world. 
With the growing number of human-AI interactions, grounding is a fundamentally important capability of AI systems, models, and agents \cite{chandu2021grounding, davidsson1993toward, pickering2004toward, ziemke1999rethinking}.  
Grounding allows AI systems to bridge semantic gaps in the real world, 
team with other agents in such environments, process inputs from the environment, and learn from interactions.
A successful synthetic teammate requires several cognitive capacities, including situation assessment, task behavior, language comprehension and generation \cite{ball2010synthetic}, and knowledge gap resolution processes. Grounding enables agents with different capabilities to communicate.

\begin{figure*}
    \centering
    \includegraphics[scale=0.45,trim={0.65cm 4cm 2.75cm 0.8cm}, clip]{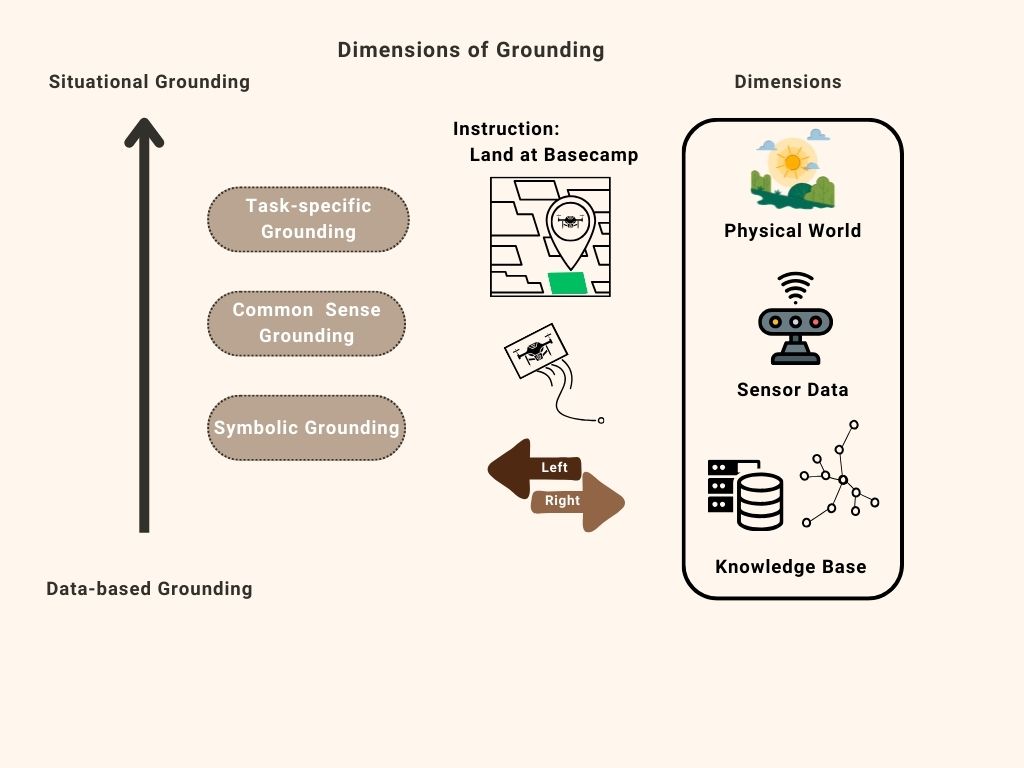}
    \caption{Grounding has multiple dimensions and needs knowledge at different levels of abstractions (this is similar to the need for linguistic, common sense, world model, and domain knowledge for language understanding; see Fig. 3 of (\url{https://bit.ly/KiLU}). Grounding may occur at different levels in the task execution process. For example, the drone agent must be able to comprehend the instructions in a natural language format. Next, the instruction may require parsing relevant symbols (in the case of neuro-symbolic methods) for reasoning processes. The information extracted must then be grounded to the drone's capabilities (common-sense grounding). Finally, the instruction must be grounded in the specific navigation task.}
    \label{fig:grounding_levels}
\end{figure*}

Both cognitive scientists and computer scientists have focused on how to make internal mechanisms (or representations) of external entities intrinsic to the agent itself rather than being defined by an external designer or interpreted by an observer \cite{ziemke1999rethinking}. 
Recent efforts in Natural Language Processing (NLP), Computer Vision (CV), and Human-Computer Interaction (HCI) improve the grounding of machine agents. However, it remains a multi-dimensional challenge, encompassing diverse contexts, abstractions, and modalities of understanding (see Figure \ref{fig:grounding_levels}). 
In the absence of a clear definition, we are unable to determine genuine advances or task-specific adjustments.

This article sheds light on different aspects of grounding through the lens of cognitive science and artificial intelligence (AI), discusses specific neuro-symbolic solutions for grounding, and highlights future work). 

\vspace{1mm}
\fbox{\begin{minipage}{\dimexpr0.8\columnwidth\fboxrule+1\fboxsep}
``A successful synthetic teammate requires several cognitive capacities including situation assessment, task behavior, language comprehension and generation \cite{ball2010synthetic}, and knowledge gap resolution processes. Grounding enables agents with different capabilities to communicate."
\end{minipage}}
\vspace{1mm}

\section*{Cognitive Science Lens}
\label{sec:grounding_cog_sci}
Identification of the symbol grounding problem in cognitive science dates to Harnad.
Following the introduction of computation with symbols credited to McCarthy, Searle's Chinese Room Problem revealed how amodal symbol manipulation lacks grounding. 
Yet, symbol manipulation is often the foundation of contemporary AI systems.  
Challenges to amodal symbol manipulation include embodied grounding \cite{nunez1999embodied} and the recent use of language and simulation to establish grounding \cite{barsalou2010grounded}. 
Given this long-standing research problem, Ziemke \cite{ziemke1999rethinking} groups grounding efforts into two categories: 1) \textit{cognitivist} or 2) \textit{enactivist}. 

\textit{Cognitivism} grounds atomic primitives in sensorimotor invariants \cite{ziemke1999rethinking}. Concepts constructed from these inherit the grounding of their constituents. 
Nevertheless, different agents may reason with different abstractions, creating a divergence that requires repair.
A more recent perspective, \textit{enactivism} values the role of action, embodiment, and environment. 
Robotic agents can potentially obtain grounding by physically linking to an environment through sensory input and motor output.
Agent functions can be either engineered or learned. With meticulously engineered grounding, systems may demonstrate the `correct' behavior, but their internal mechanisms are not inherent to the system.
Alternatively, an agent function can be acquired by adjusting connection weights instead of requiring programming. 
However, the definition of a correct agent function and how to evaluate various agent functions remains a challenge.

Given our interest in real-world human-AI interaction and multimodal systems, we
emphasize enactivist grounding. However, cognitivist grounding remains essential, as we summarize from Barsalou \cite{barsalou2010grounded}:

\begin{quote}
Mental imagery, cognitive grammar, mental spaces, and compositional reasoning support explanations for thought. Neuroimaging shows that higher cognition is realized in the brain's \textit{model systems}. One theory is that \textit{grounding mechanisms serve as an interface}, peripheral to core cognitive operations.
\end{quote}

Barsalou predicts that future cognitive research will integrate classic symbolic architecture, statistical/dynamic systems, and grounding cognition. 
Formal and computational accounts of grounding will shift from epiphenomenal to casual. 
Grounding mechanisms may potentially replace the amodal mechanisms in cognitive architectures.
We believe that advancements in neuro-symbolic AI will be instrumental here.
Next, we describe the grounding problem from a machine learning and AI perspective and highlight what is missing from a cognitive science lens. 

\section*{Artificial Intelligence Lens}
\label{sec:grounding_ai}
Grounding in AI has typically been referred to as connecting concepts to other knowledge bases or world models. We first discuss the different grounding efforts in the AI~\cite{chandu2021grounding} community. We then draw parallels with the cognitive science community\cite{ziemke1999rethinking, barsalou2010grounded} and identify some nuanced distinctions.

\begin{figure*}
    \begin{center}
    \begin{subfigure}{2\columnwidth}
    \centering
        \includegraphics[scale=0.6, trim={1.65cm 10.75cm 2.75cm 2.25cm}, clip]{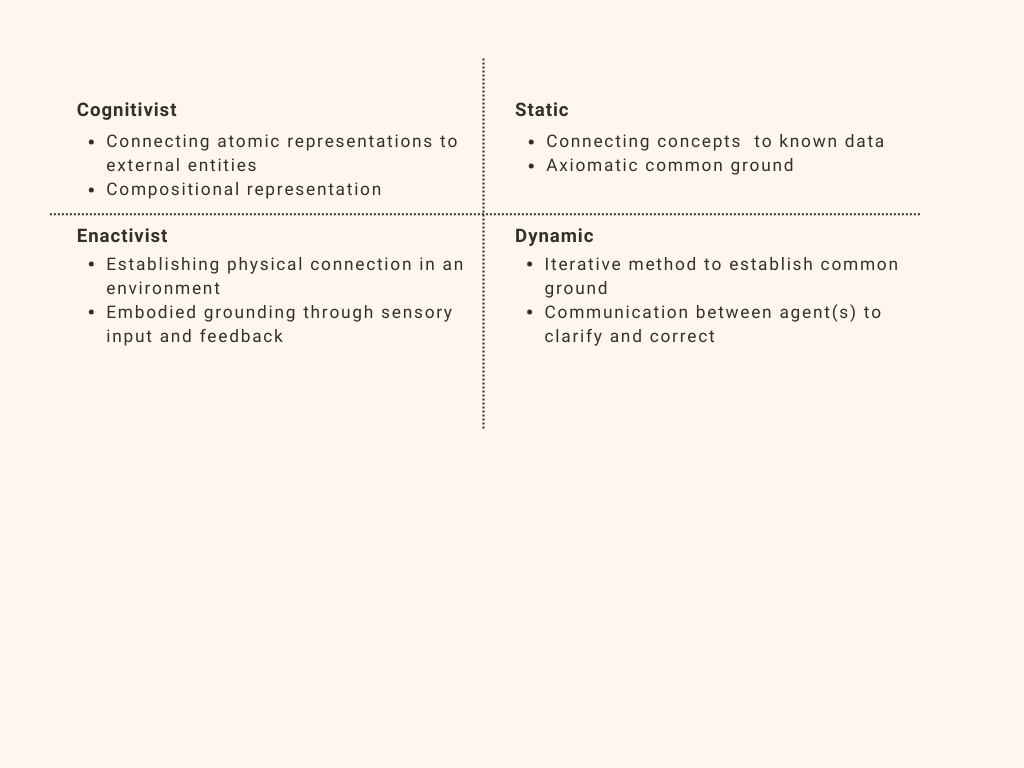}
    \caption{
The \textit{static} and \textit{dynamic} definitions in neurosymbolic AI loosely mirror the \textit{cognitivist} and \textit{enactivist} grounding in cognitive science, respectively, although there are subtle differences. 
Cognitive scientists tend to focus more on the specific monitoring and repair processes in case of different perspectives between interacting agents, whether due to differences in static knowledge or engagement opportunities with an open world.}
\end{subfigure}%
\end{center}

    \begin{subfigure}{2\columnwidth}
        \centering
        \includegraphics[scale=0.5, trim={0.20cm 2.5cm 3.45cm 0.45cm}
, clip]{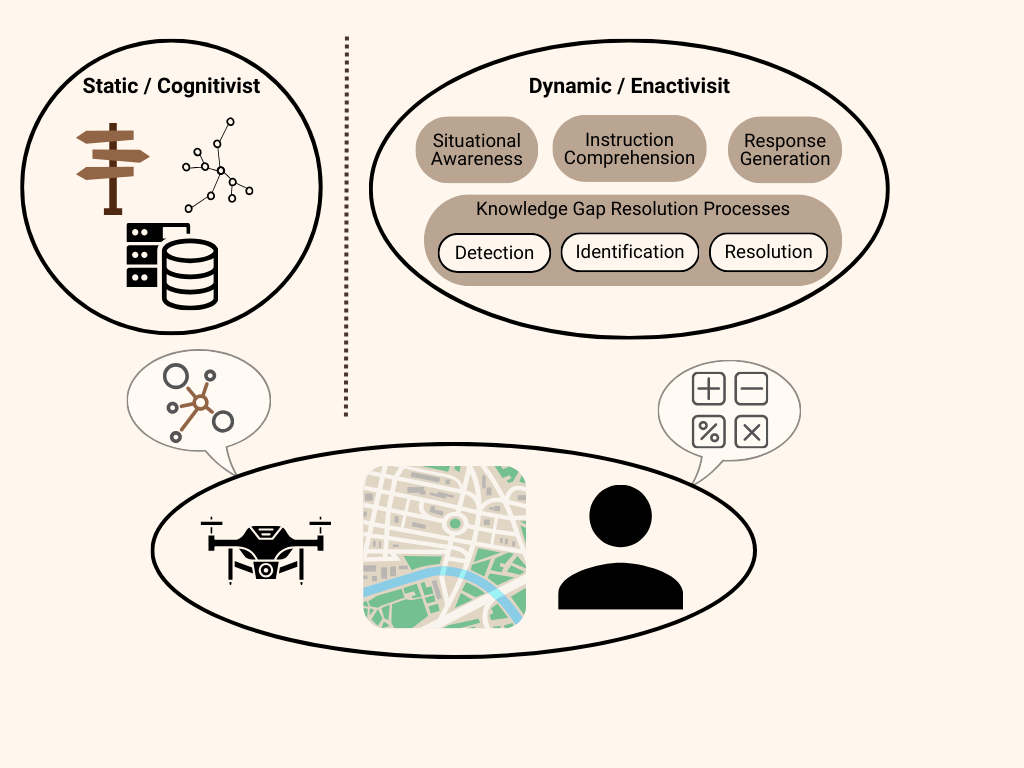}
\centering
\caption{
For an AI or cognitive agent to interact effectively with its environment, it must understand the language used by external agents, accurately assess the current situation and context, and identify and address gaps in its knowledge \cite{bajaj2022knowledge}. This process involves recognizing when it lacks understanding or has difficulty with language comprehension and communicating these knowledge gaps to other agents or external sources. By utilizing these processes, agents can update their knowledge of the task and environment, avoiding catastrophic errors and allowing for the iterative and interactive aspects of dynamic grounding. Consider a machine agent, like a drone, working with a human operator to survey a specific area. The agent can use its background (stored) knowledge to understand the monitored environment and the mission's purpose (static/cognitivist grounding). Suppose the agent comes across something it has not seen before and cannot identify. In that case, it will interactively and iteratively ask the human operator for assistance in closing this gap (dynamic/enactivist grounding).
}
    \end{subfigure}
    \caption{Types of Grounding}
    \label{fig:s_d_grounding}
\end{figure*}

Chandu et al.~\cite{chandu2021grounding}, note the use of \textit{static} versus \textit{dynamic} terminology used in the NLP / CV context,  similar to the \textit{cognitivist} versus \textit{enactivist} distinction in cognitive science (see Figure \ref{fig:s_d_grounding}). Static grounding -- the most predominant form, relies on accessible evidence supporting the \textit{common ground} to connect concepts within a given context to the real world \cite{chandu2021grounding}. 
For static grounding, common ground is typically through an agent (e.g., machine) interacting with a static knowledge base (KB) to frame a response and deliver it back to the other agent (e.g., human). 
Both agents may share this common ground by assuming its universality, i.e., no external references. The success of grounding is measured based on the agent's ability to link the query to the available data. 
In contrast, dynamic grounding establishes common ground iteratively, where both agents can communicate to seek and provide clarifications, typically in a potentially changing physical environment. 
This allows corrections to misunderstandings.

\noindent
\textbf{Static grounding:} According to Harnad, manipulating symbolic representations without meaning cannot support reasoning. 
AI researchers responded with different representations for different uses. 
Typically, these frameworks have the following key components: the \textit{designator}, denoting the name or symbol utilized to identify the category; the \textit{epistemological representation}, employed to recognize instances of the category; and the \textit{inferential representation}, comprising ``encyclopedic" knowledge about the category and its members. 
The \textit{epistemological representations} are termed \textit{concept descriptions}. In computer vision, these representations are considered object models \cite{davidsson1993toward}. 
This form of grounding is typically established using deductive learning. It is consistent with cognitivist grounding, with little to no active or online supervision from the environment or an external agent.  
ML efforts for static grounding include using entity slot filling, adversarial references to grounding visual referring expressions, visual semantic role labeling, and disambiguation of concepts and entities \cite{chandu2021grounding}. 
Additionally, methods designed for manipulating representations include fusion and concatenation, representation alignment, and projection of representations into a shared space. 
Finally, the ML community has designed different learning objectives to address the grounding problem, including multitasking and joint training, the design of new loss functions, and adversarial learning methods. 

\noindent
\textbf{Dynamic grounding:} Dynamic efforts in ML are typically designed for situations where an entity in an environment is matched with an epistemological representation that activates a larger knowledge structure containing the composite concept representation. Such systems learn to  ground their own experience dynamically in the environment, creating more robust capabilities not dependent on pre-programmed representations \cite{davidsson1993toward}. 
Grounding frameworks such as \textit{learning from example and learning by conversation} are consistent with enactivist grounding in cognitive science. Efforts in the ML community for dynamic grounding include grounding embodied agents, natural interactions with human-in-the-loop feedback, and, more recently, grounding LLM-based agents (\url{https://rb.gy/2pfq1g}).  
Nevertheless, grounding in mainstream machine learning exploits \textit{deductive learning} \cite{davidsson1993toward}. 

\noindent
\textbf{Limitations:} One notable omission in ML is grounding with sensors or environmental data \cite{cangelosi2006embodied}. 
Furthermore, the study of \textit{latent pragmatics} is also missing from Chandu et al.'s work. Pragmatic analysis, pioneered by Austin, Grice, and Searle and extended by others such as Sperber, focuses on understanding functional intentions and implications based on variations in linguistic content 
across different contexts \cite{searle1969speech, leech2016principles, jurafsky2006pragmatics}.

The lack of a consistent definition of grounding creates considerable ambiguity regarding \textit{how to ground, what to ground, and where to ground}. To bridge this gap, Chandu et al. \cite {chandu2021grounding}
outline the following grounding stages: Stage 1 - Localization, Stage 2 - External Knowledge, Stage 3 - Common Sense, and Stage 4 - Personalized Consensus. However, these stages are insufficient.
Consider a challenging example of grounding a drone (with an AI-based decision-making model) to ensure safety concerns.  
Localization requires that the drone accurately determine its position relative to its surroundings. 
Next, grounding with external knowledge introduces additional information, such as weather and airspace regulations, 
that inform subsequent action.
However, which external knowledge sources should be used and prioritized in the decision-making?
Stage 3 - Common sense could include avoiding obstacles in its flight path and taking proactive measures to avoid risks or mitigate potential harm. However, it is not easy to quantify and measure this grounding stage. 
Stage 4 - Personalized consensus can help ground decisions to the drone's perception of the environment and prior experiences.
However, it is unclear whether the drone must adhere to its prior experiences when it conflicts with human instructions that reflect a broader understanding of the situation.
While a helpful initial attempt, these stages are not sufficiently specific to standardize how grounding occurs and at what abstraction levels.
The lack of definition allows for creative interpretation of the problem and, therefore, new tasks, datasets, and methods. 
However, in the future, the community will benefit from a standardized, comprehensive definition. 

\vspace{1mm}
\fbox{\begin{minipage}{\dimexpr0.8\columnwidth\fboxrule+1\fboxsep}
``The lack of a consistent definition of grounding creates considerable ambiguity regarding \textit{how to ground, what to ground, and where to ground}."
\end{minipage}}
\vspace{1mm}

\section*{Neuro-Symbolic Grounding}
\label{sec:ns_for_grounding}
Neuro-symbolic methods can benefit grounding by integrating traditional symbolic reasoning approaches with the generalization capabilities of neural networks. Neuro-symbolic systems integrate the statistical learning capabilities of neural networks with the structured, symbolic representations used in classical AI.
Neuro-symbolic AI seeks to benefit from the synergy of symbolic and neural methods. 
Traditional symbolic reasoning methods use formal languages (e.g., Planning Domain Definition Language (PDDL)) for reasoning over knowledge stored in a structured format and represented by `symbols.' These reasoning methods manipulate and infer from structured symbolic representations, such as logic-based rules, knowledge graphs, or ontologies. The symbolic representations provide a transparent and interpretable framework for knowledge representations and logical reasoning. However, these systems are brittle and often cannot be generalized. 
But combined with neural methods,  traditional methods leverage advancements in deep learning to acquire knowledge representations and enhanced generalization capabilities effectively \cite{sheth2023neurosymbolic}. Neural networks consist of interconnected layers of \textit{artificial} neurons that use weighted connections, enabling them to learn complex mappings between inputs and outputs. These networks are instrumental in pattern recognition, classification, regression, and sequence prediction. Therefore, neuro-symbolic methods leverage the strengths of each paradigm.

As noted above, many of the grounding efforts in ML rely on deductive learning and lack active or online supervision from the environment or an external agent. Neuro-symbolic methods for grounding can offer several advantages, including compositional reasoning and situational awareness. 

\vspace{1mm}
\fbox{\begin{minipage}{\dimexpr0.8\columnwidth\fboxrule+1\fboxsep}
``Neuro-symbolic methods for grounding can offer several advantages, including compositional reasoning and situational awareness."
\end{minipage}}
\vspace{1mm}

One particular asset of neuro-symbolic methods is the use of functional modules. 
Natural language texts (i.e., instructions or queries) are mapped to functional modules that carry out atomic actions. 
These functional modules can be user-defined or learned. 
This allows agent functions grounded in symbolic representations to complete specific actions or generate responses. 
Additionally, the compositional nature of these functional programs allows for generalization to new combinations of parsed instructions or queries. 
The functional modules can be used for both \textit{dynamic} and \textit{static} grounding as the modules can operate over knowledge bases (including KGs). 
This aspect of neuro-symbolic methods can help establish common ground, enabling agents to interpret and execute instructions in a manner that aligns with symbolic human reasoning. 
For example, Mao et al. \cite{mao2019neurosymbolic} designed a new neuro-symbolic concept learner that can learn embeddings for symbolic visual inputs. Learning these mappings allows for continual learning and adaptation of new environmental variables while preserving an agent's task behavior. 
In this manner, neuro-symbolic approaches can allow for the grounding of new concepts in an environment and facilitate knowledge gap detection, identification, and resolution processes leading to adaptive and robust models. 

\section*{Concluding Remarks}
\label{sec:future_work_conclusion}

We have focused on explicit notions of grounding. We conclude 
with remarks on implicit forms of grounding.
Recently, the use of digital twins has emerged to augment the performance of machine learning systems in domains ranging from autonomous transportation \cite{mcclellan2022physics} to next-generation wireless communication \cite{banerjee2024accurate}. Digital twins can provide a high-fidelity representation of physical entities by accurately modeling the structure, behavior, and characteristics of a real-world system (world model) governed by physical laws. Such ideas can expand the training dataset (out-of-distribution), allowing models to be generalized.  
This implicit grounding in physical laws can prevent non-factual generation, reduce hallucinations, and anchor model responses to specific information, facilitating harmonization with the corresponding world model. 
We note that using digital twins for grounding parallels the contention of Pickering and Garrod \cite{pickering2004toward} that interlocutors implicitly comprehend each other by aligning their models of the discussed situation at various levels of cognitive and linguistic representation. 
Such implicit alignment processes between agents (akin to digital twins) inspire computational grounding processes.

Finally, we advocate for more knowledge-infused neuro-symbolic learning and reasoning systems that naturally integrate linguistic, common-sense, general (world model), and domain-specific processes and knowledge to facilitate static grounding. We expect fundamental progress on the synergistic use of neural networks and structured semantics to advance from content processing to content understanding and reasoning with the infusion of symbolic knowledge (\url{https://rb.gy/67wgx3}). These methods require dynamic knowledge-elicitation strategies integrating multimodal pragmatic context and interactions with domain experts in the loop to achieve dynamic grounding and alignment with user intent. 

\smallskip
\noindent {\textbf{Acknowledgements}:}
We acknowledge the National Science Foundation (NSF) Award \#2335967 (AS and VS) and 
 NSF award \#CNS-2112471 (GB and SP).





\input{bib}

\end{document}

%% file: bib.tex
\bigskip
\noindent
{\textbf{Goonmeet Bajaj}} is a Ph.D. Candidate at the Ohio State University (OSU) and is interested in cognitive-inspired AI. Contact her at bajaj.32@osu.edu.

\noindent
{\textbf{Srinivasan Parthasarathy}} is a Professor of Computer Science \& Engineering and co-directs the Data Analytics program at OSU. He is a fellow of the IEEE, AAIA and the Robert Bosch Center for Data Science \& AI.  Contact him at srini@cse.ohio-state.edu.


\noindent
{\textbf{Valerie Shalin}} is a Professor of Psychology at Wright State University. Contact her at valerie.shalin@wright.edu.

\noindent
{\textbf{Amit Sheth}} is the founding director of the AI Institute, NCR Professor of Comp. Sc.\& Engg. at USC. He received the 2023 IEEE-CS Wallace McDowell award and is a fellow of IEEE, AAAI, AAIA, AAAS, and ACM. Contact him at amit@sc.edu.